\documentclass{article}

\usepackage{arxiv}

\usepackage[utf8]{inputenc} % allow utf-8 input
\usepackage[T1]{fontenc}    % use 8-bit T1 fonts
\usepackage{hyperref}       % hyperlinks
\usepackage{url}            % simple URL typesetting
\usepackage{booktabs}       % professional-quality tables
\usepackage{amsfonts}       % blackboard math symbols
\usepackage{nicefrac}       % compact symbols for 1/2, etc.
\usepackage{microtype}      % microtypography
\usepackage{lipsum}		% Can be removed after putting your text content
\usepackage{graphicx}
\usepackage{natbib}
\usepackage{doi}

\title{Residual-Concatenate Neural Network with Deep Regularization Layers for Binary Classification}

\author{
    Abhishek Gupta\\
	Research Scholar\\
	University of Mumbai\\
	Mumbai, MH, India \\
	\texttt{abhishek.gupta20001@gmail.com} \\
	\And
	Sruthi Nair \\
	Master of Engineering \\
	Vidyalankar Institute of Technology\\
	Mumbai, MH, India \\
	\texttt{sruthi.rk.nair@gmail.com} \\
	\AND
	Raunak Joshi \\
	Mentor \\
	University of Mumbai \\
	Mumbai, MH, India \\
	\texttt{raunakjoshi.m@gmail.com} \\
	\And
	Vidya Chitre \\
	Assistant Professor - Department of IT \\
	Vidyalankar Institute of Technology\\
	Mumbai, MH, India \\
	\texttt{vidya.chitre@vit.edu.in} \\
}

\renewcommand{\shorttitle}{\textit{Residual-Concatenate Neural Network with Deep Regularization Layers for Binary Classification}}

\hypersetup{
pdftitle={Residual-Concatenate Neural Network with Deep Regularization Layers for Binary Classification},
pdfsubject={cs.LG},
pdfauthor={Abhishek Gupta, Sruthi Nair, Raunak Joshi, Vidya Chitre},
pdfkeywords={Concatenation, Regularization, Residuals},
}

\begin{document}
\maketitle

\begin{abstract}
	Many complex Deep Learning models are used with different variations for various prognostication tasks. The higher learning parameters not necessarily ensure great accuracy. This can be solved by considering changes in very deep models with many regularization based techniques. In this paper we train a deep neural network that uses many regularization layers with residual and concatenation process for best fit with Polycystic Ovary Syndrome Diagnosis prognostication. The network was built with improvements from every step of failure to meet the needs of the data and achieves an accuracy of 99.3\% seamlessly.
\end{abstract}

% keywords can be removed

\section{Introduction}
Deep Learning\citep{lecun2015deep} definitely has made many advancements in last decade by introducing many different types of neural networks. These include solving different problems from the domain of vision, text and audio. The area of deep learning has been using neural networks which is a learning procedure that replicates the biological human neuron system. It learns representations and patters from the data like humans do. These representations are highly adaptable with changes in the learning patterns to best fit the needs of the problem. The standard neural network has an input layer and an output layer with $n$ number of hidden layers. These hidden layers are responsible for the learning process that takes place and gives to the output layer. The entire process of any neural network is a two-step process consisting of forward propagation\citep{548917} and backward propagation\citep{Rumelhart1986LearningRB}. The forward propagation is responsible for learning patterns from the input data and backward propagation is responsible for improving the losses incurred in the learning process of forward propagation. The input layer considers the data in form of tensor and initializes the weight and bias for the network. These parameters are passed later to the hidden layers. These hidden layers are responsible for modulating these weight and biases with respect to input tensor and helps learn representations. These weight and biases alone are not effective without the use of activation functions\citep{nwankpa2018activation} which are a set of mathematical functions that purely focus on the aspect of considering the right aspect of information from the entire representation. These activation functions differ from task to task, considering binary classification\citep{10.5120/ijca2017913083}, an activation function like sigmoid\citep{10.1016/S0893-6080(05)80129-7} will be used that differentiates the 2 classes very effectively. Assuming classes in the labels are more than 2, softmax\citep{NIPS1989_0336dcba}\citep{sep-statphys-Boltzmann} is used which gives probabilities of all the predictable classes where the class with the highest probability can be extracted very efficiently. For learning representations from the set of hidden layers that are interconnected an activation function like tanh\citep{Ryck2021OnTA} or non-linearity activation like rectified linear unit also known as ReLU\citep{agarap2018deep} is used. Amidst this learning process the loss is calculated with a loss based function. These loss functions compute and keep track of the information lost during learning representations. This completes the process of forward propagation and process is transferred to backward propagation where the gradients for loss are computed and changes in learnt weights and biases are computed. This entire process consists of one epoch\citep{Afaq2020SignificanceOE}. Now in this paper we focus particularly on binary classification task with a deep neural network. When the number of hidden layers are very high in number the neural network is considered as deep neural network. Now training a very deep neural network can run into a problem of learning representations either more or less. This is where regularization techniques are introduced and these prove to be a better for improving the performance. Considering the binary classification problem, we had a wide horizon for selecting data. In this case we considered working with Polycystic Ovary Syndrome diagnosis dataset. This is disorder observed in women and the data is very high dimensional and conducive for binary classification. Machine Learning is widely used for PCOS prognostication in many different contexts. Considering most commonly used algorithms of linear models or distance based algorithms\citep{9510128} has been done. Apart from that work with Discriminant Analysis\citep{gupta2022discriminant} is done. Extending to ensemble learning\citep{8929674} and the varieties it provides, the work with bagging\citep{kanvinde2022binary}, boosting\citep{gupta2021succinct} and stacking\citep{nair2022combining} has been performed effectively. In this paper we specifically target the weaker sections of all state-of-the-art approaches and try to come up with techniques to get past the transgressions of learning methods. 

\section{Methodology}
\subsection{Basic Network}
The basic network is build to have a base to start with for making changes. The data used has 41 dimensions of feature set and 1-dimensional vector of labels. Training of 3-5 hidden layers is done specifically that targets the basic learning with one output layer with 1 hidden neuron and sigmoid activation function. The hidden neurons from hidden layers are highly arbitrary and loss function used is binary cross-entropy with Adam\citep{kingma2017adam} loss optimization\citep{reddy2018optimization} function. Now this network is particularly not efficient for learning representations and runs into an under-fitting\citep{8927876} problem. So this is where a more deeper network needs to be trained so the number of hidden layers will be increased but the model will give high variance\citep{Mehta2019AHL} and run into an over-fitting\citep{8927876} problem. For this the need of regularization arises.

\subsection{Weight Decay}
\begin{figure}[htbp]
    \centering
    \includegraphics{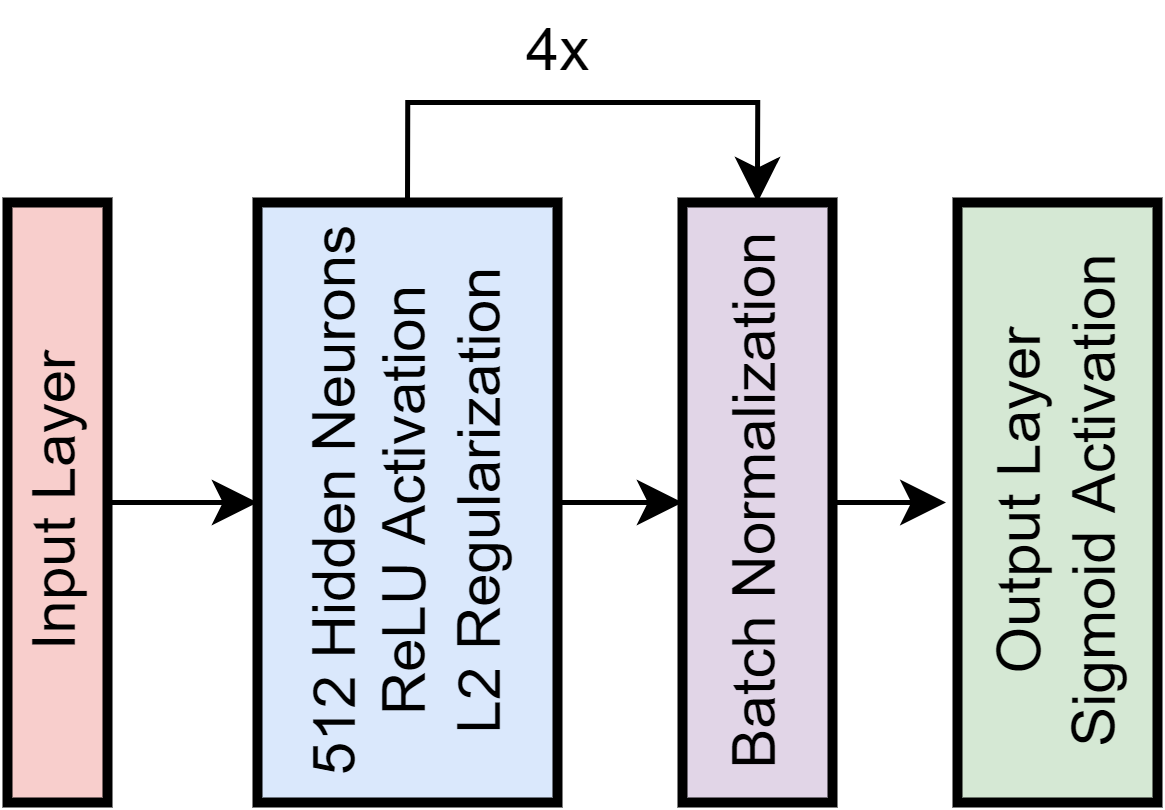}
    \caption{L2 Regularization with Batch Normalization}
    \label{fig:fig1}
\end{figure}

The weight decay\citep{xie2021understanding} process is also known as regularization and there are 2 types, $L_1$ and $L_2$\citep{Ng2004FeatureSL}\citep{szabo2004l1} regularization and it directly works for solving the over-fitting problem of the networks. Considering the $L_2$ regularization first, the regularization parameter is Euclidean Norm\citep{wu2019deep} and can be defined by the sign $\Omega$. It considers the weight matrix of the learning representation and computes sum over all the squared weight values of it. The trained network can be seen in the Fig. \ref{fig:fig1}

The batch normalization layer has non-learnable parameters that cuts off the excess slack for any neural network learning process very efficiently. The formula for the $L_2$ Regularization is given as

\begin{equation}
    \Omega(W) = \sum_i \sum_j W^2_{ij}
\end{equation}

where the W is the weight and now this further changes the loss function. One scalar parameter of $\alpha$ is used. This changes the loss function and can be given as

\begin{equation}
    L_2(W) = \frac{\alpha}{2} \sum_i \sum_j W^2_{ij} + L(W)
\end{equation}

where the $L_2$ is regularized loss function and along with the Euclidean Norm the loss function is added and changes are made. Similarly same architecture with $L_1$ Regularization is used. It considers the absolute values of the weight parameters. The formula for this can be given as

\begin{equation}
    \Omega(W) = \sum_i \sum_j |W_{ij}| \nonumber
\end{equation}

and loss function can be updated as

\begin{equation}
    L_1(W) = \alpha \sum_i \sum_j |W_{ij}| + L(W)
\end{equation}

Now these architectures are not performing efficiently and the changes can be made using a different procedure.

\subsection{Concatenation}
This is a very different process where you combine best of both networks. In this the network are trained in a parallel style. Assume training multiple networks simultaneously and the outputs are received at other end where they are combined and given to the output layer. This has an advantage as well as disadvantage. The advantage is that results can be improved as more parameters are learnt by network. Disadvantage can be that if one network is not able to perform effectively, the results can be impacted. But this procedure proves to be effective in most of the cases.

\begin{figure}[htbp]
    \centering
    \includegraphics[scale=0.8]{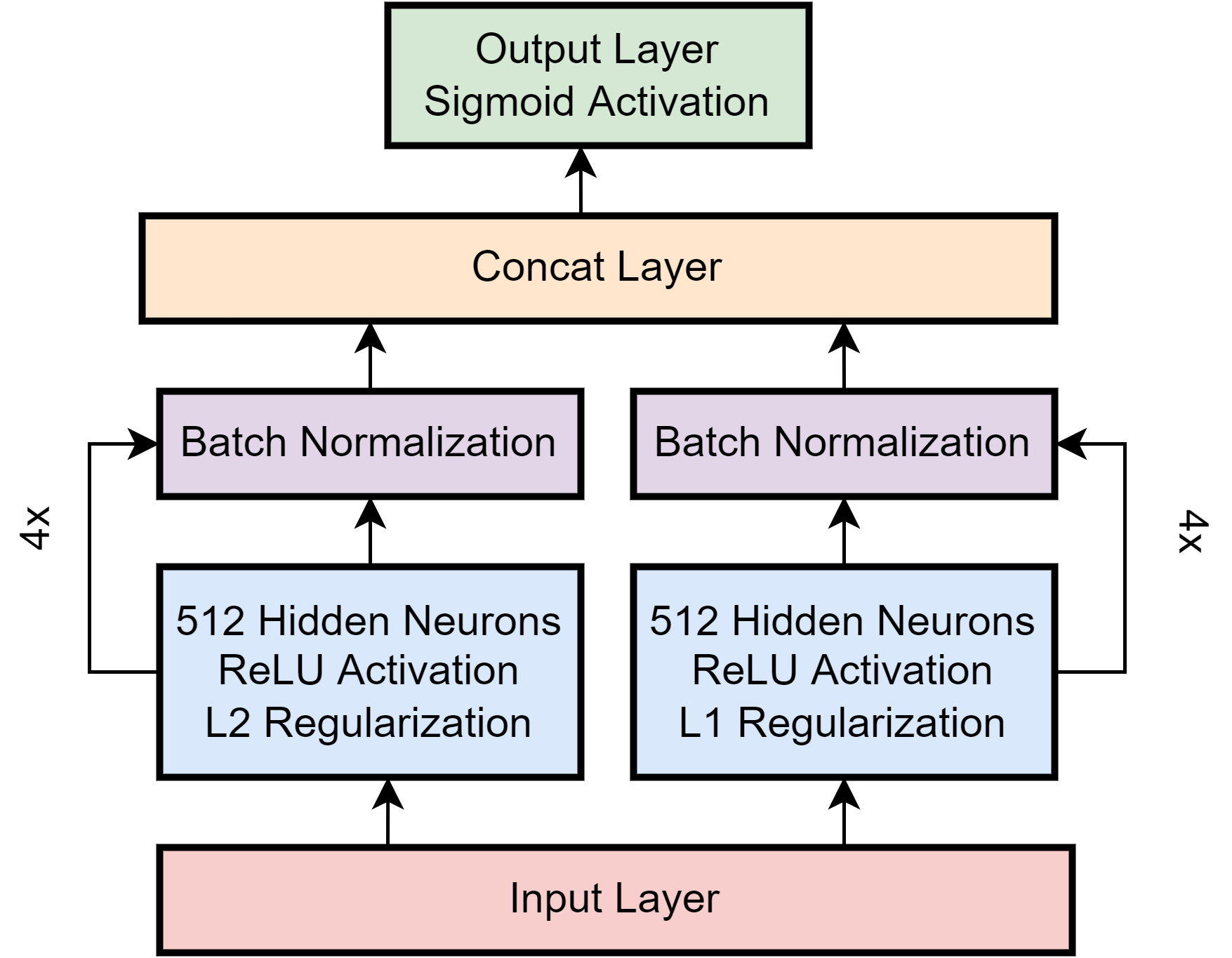}
    \caption{Regularization Model with Concatenation}
    \label{fig:fig2}
\end{figure}

The Fig. \ref{fig:fig2} is the graphical representation of the model. The input simultaneously imparts the data to 2 networks independently which train in the parallel fashion and give their respective outputs to the concatenation\citep{Noreen2020ADL} layer. The concatenation layer has the rule that the dimensions of both the networks are supposed to be same. We can consider the number of hidden neurons in layman terms. If they are not same the operation is not performed effectively. Later the output from the concatenation layer is given to the output layer. This process increases the number of trainable parameters. With this type of network we are able to curb the under-fitting problem but now the network runs into an over-fitting problem. Since the network trained is very deep and fits on the data very efficiently and does not consider the validation data effectively, more new variation needs to be introduced.

\subsection{Residuals}
\begin{figure}[htbp]
    \centering
    \includegraphics[scale=0.75,angle=90]{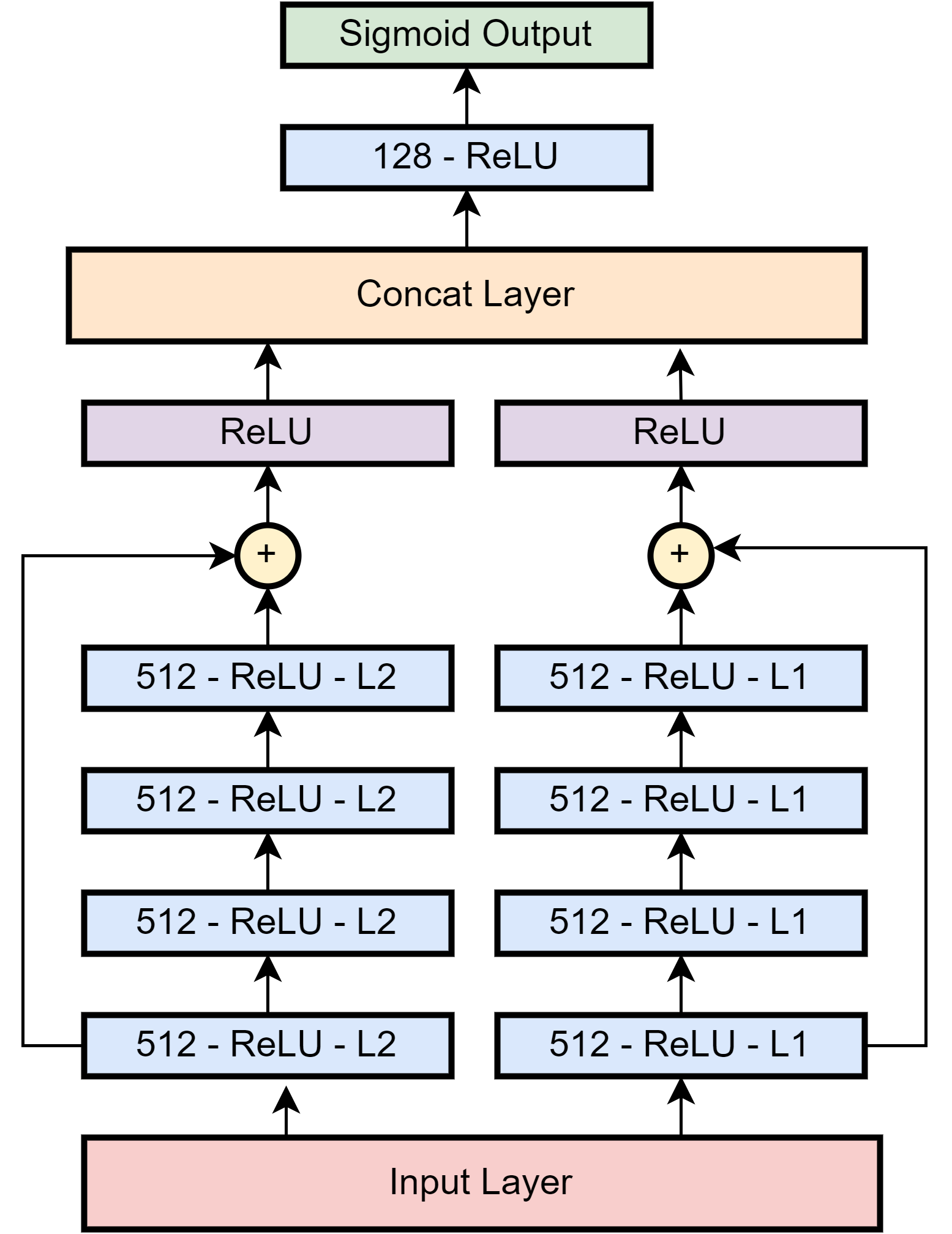}
    \caption{Residual Concatenated Model}
    \label{fig:fig3}
\end{figure}

Right before understanding the concept of Residual, the final architecture can be represented with Fig. \ref{fig:fig3}

The residual\citep{he2016deep} blocks are the main essence of this architecture. The over-fitting issue persists because of the saturation in the learning process. Deeper the layers, more representations will be computed and the calculating the gradients in the back propagation becomes more complex. The residual block inspired by ResNet\citep{he2016deep}, works with the concept of skip connections or residual connections. The skip connections\citep{wu2020skip} particularly work when the degradation problem starts. The batch normalization\citep{ioffe2015batch} layer was eliminated for the fact that network was over-fitting. Highway networks\citep{srivastava2015highway} was the concept from which the residual blocks were derived. The calculable parameters are important where the $L_1$ model has 809472 parameters, whereas $L_2$ model has the same parameters. The total amount of parameters of combined network are 1750273 parameters. The activation functions used in the hidden layer are ReLU\citep{agarap2018deep}. The number of hidden neurons are 512 in every single layer. The output from both the regularization models is applied with a residual block and the output is applied with a ReLU layer. Later the output from ReLU layer is given to the concatenation layer. The output of the concatenation layer is given to a 128 hidden neuron layer with ReLU activation. This finally gives output to the last layer with sigmoid activation function.

\section{Results}
The main inferences can be derived from the result section. The results consider results of all the different models used and what was the need for using a better model. 

\begin{figure}[htbp]
    \centering
    \includegraphics[scale=0.55]{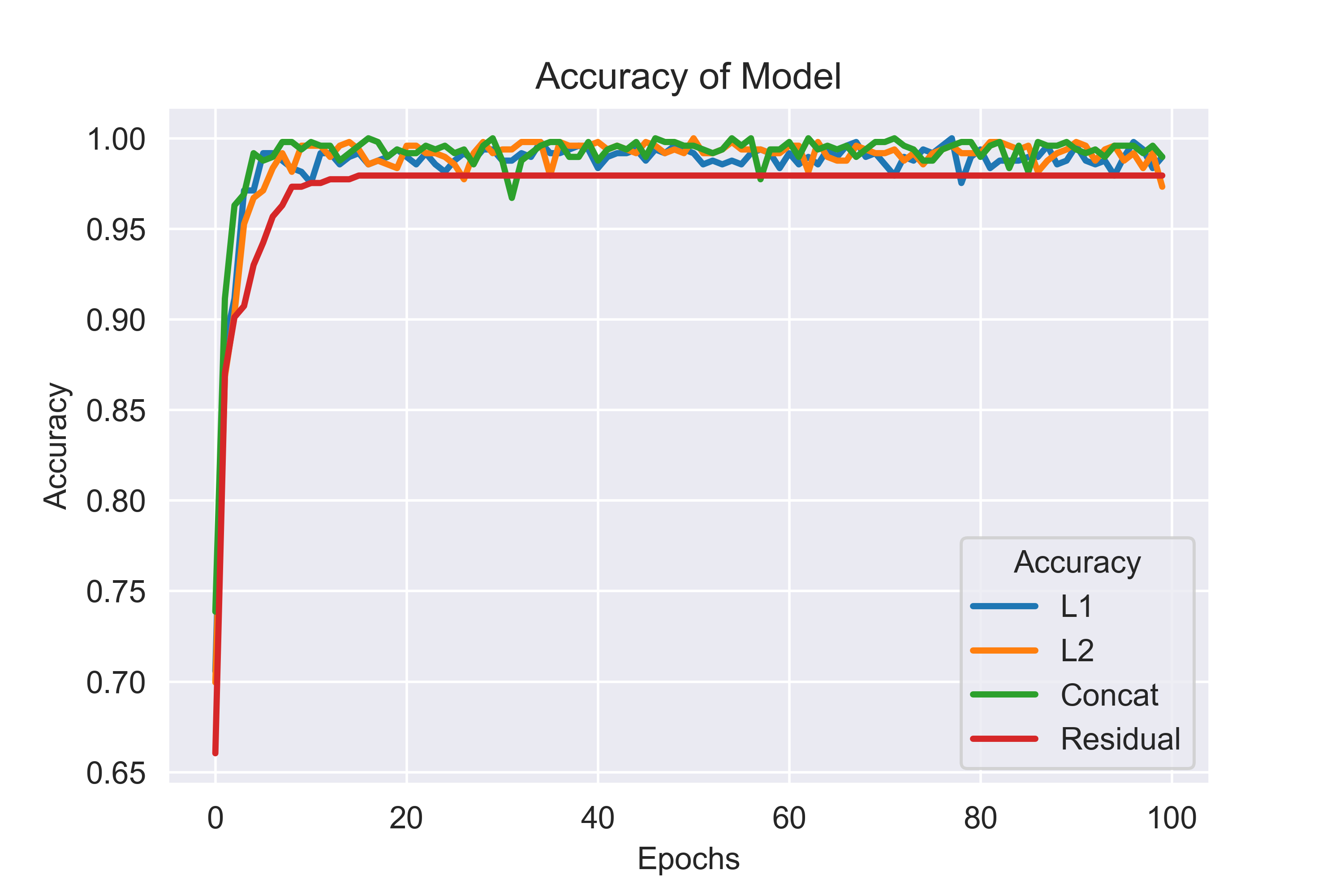}
    \caption{Training Accuracies}
    \label{fig:fig4}
\end{figure}

The accuracy and loss can be visualized to have an intuition of how the performance of the model is taking place. Similarly the graph for loss can also be visualized that shows the loss reduced with respect to number of epochs with all the model variations.

\begin{figure}[htbp]
    \centering
    \includegraphics[scale=0.55]{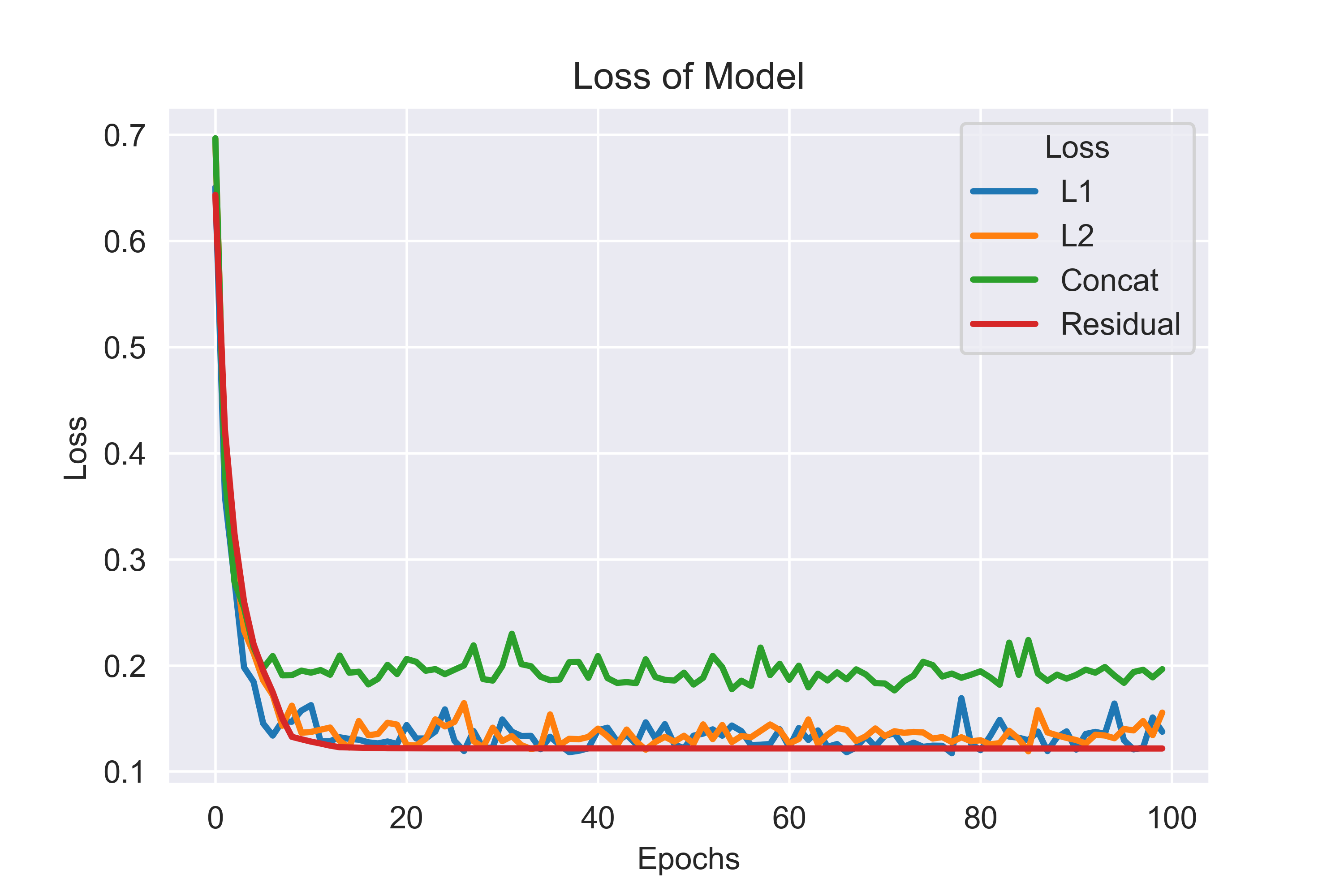}
    \caption{Training Losses}
    \label{fig:fig5}
\end{figure}

The Fig. \ref{fig:fig4} and \ref{fig:fig5} give very distinct visual representations of the model training performance, though the question arises how come we assume the residual model performs better even though the training accuracy is lower as compared to other models. The answer to this is the differences between training and validation accuracy. It is a key determining factor for defining the state of the model. If the training accuracy is way lower than expected, it is said to be in state of under-fitting, while if accuracy of training is way higher and testing accuracy is pretty much stagnant, inference can be drawn that model is in state of over-fitting. The accuracies of the other models is 100\% and convergence of the loss is not as good as expected. This is the need for the residual model.

\begin{table}[htbp]
\caption{Training and Validation Accuracies}
\begin{center}
\begin{tabular}{lll}
\hline
Algorithm & Training Accuracy & Validation Accuracy \\
\hline
$L_1$-Reg & 100\% & 92.72\%  \\
$L_2$-Reg & 100\% & 94.54\%  \\
Concat & 100\% & 92.72\% \\
Residual & 99.3\% & 95.4\% \\
\hline
\end{tabular}
\label{tab:tab1}
\end{center}
\end{table}

The Table \ref{tab:tab1} gives the proof for the need of the Residual changes. The other models are over-fitting and residual block network did performed better than the others. This pretty much eliminated the need of observing the values of the loss table. But just for the sake of deriving the perfect inference, let us look at Table \ref{tab:tab2}.

\begin{table}[htbp]
\caption{Training and Validation Losses}
\begin{center}
\begin{tabular}{lll}
\hline
Algorithm & Training Loss & Validation Loss \\
\hline
$L_1$-Reg & 11.91\% & 28.81\%  \\
$L_2$-Reg & 11.73\% & 25.9\%  \\
Concat & 17.65\% & 36.13\% \\
Residual & 8.11\% & 21.48\% \\
\hline
\end{tabular}
\label{tab:tab2}
\end{center}
\end{table}

When we consider the losses, residual performs the best amongst others. The need for residual was mandatory based on the performance metrics of other variations of neural network.

\section{Conclusion}
The major advancements in the area of deep learning were referred to create a very subtle architecture that points out some transgressions with binary classification problem. The demonstration of the approach was done starting from very basic of the neural networks. The data has 41 dimensional feature set. In the start a very simple 3-5 layered architecture was designed which was increased and regularization was included. $L_1$ and $L_2$ regularization was introduced that has some issues which were later addressed with concatenation model. This model has high number of parameters as it is combined to give the best of 2 worlds. Yet some over-fitting issues were experienced for which residual based blocks were introduced that gave the best performance. This paper targets a very subtle implementation of neural networks for high dimensional binary classification problem. In future by referring this paper, many new methods can be formulated for which we would be more than happy.

\section*{Acknowledgment}
Our team thanks Mr. Prasoon Kottarathil for his effort in gathering and providing the dataset of Polycystic Ovary Syndrome on Kaggle.

\bibliographystyle{unsrtnat}
\bibliography{references}

\end{document}